\PassOptionsToPackage{dvipsnames}{xcolor}
\documentclass[final,onefignum,onetabnum]{siamart190516}

\usepackage{amsfonts, mathtools, amssymb, amsopn, bm}
\usepackage{mathrsfs}
\usepackage{algorithmic}
\usepackage{booktabs, siunitx}
\usepackage{tikz, pgfplots, graphicx}
\pgfplotsset{compat=1.13}
\usepackage{epstopdf}
\usepackage[caption=false]{subfig} % <- Preamble
\usepackage{todonotes}

\DeclareMathOperator*{\argmin}{arg\,min}

\DeclareMathOperator*{\supp}{supp}

\newcommand{\fun}[3]{#1 \colon #2 \rightarrow #3}
\newcommand{\abs}[1]{\left|#1\right|}
\newcommand{\norm}[2]{\left\| #1 \right\|_{#2}}

\renewcommand{\sp}[2]{\langle #1, #2 \rangle}

\renewcommand{\subset}{\subseteq}
\renewcommand{\epsilon}{\varepsilon}
\renewcommand{\b}{\bm}
\renewcommand{\cref}{\Cref}

\newcommand{\eps}{\mathrm\varepsilon}
\newcommand{\e}{\mathrm e}

\renewcommand{\i}{\mathrm i}

\newcommand{\C}{\mathbb C}
\newcommand{\N}{\mathbb N}
\newcommand{\R}{\mathbb R}
\newcommand{\Z}{\mathbb Z}
\newcommand{\T}{\mathbb T}

\newcommand{\dom}{\mathrm{D}}
\newcommand{\tar}{\mathbb{K}}
\newcommand{\X}{\mathcal{X}}

\newcommand{\x}{\b x}
\newcommand{\y}{\b y}
\renewcommand{\k}{\b k}

\renewcommand{\d}{\,\mathrm{d}}

\renewcommand{\L}{\mathrm{L}}

\newcommand{\D}{\mathcal D}
\renewcommand{\u}{\b u}
\newcommand{\uc}{\b{u}^{\mathrm c}}
\newcommand{\au}{\abs{\u}}

\renewcommand{\v}{\bm v}

\newcommand{\av}{\abs{\v}}

\newcommand{\va}[1]{\sigma^2(#1)} 
\newcommand{\gsi}[2]{\varrho(#1,#2)}

\newcommand{\fc}[2]{\mathrm{c}_{#1}\!\left(#2\right)}

\newsiamremark{remark}{Remark}
\newsiamremark{hypothesis}{Hypothesis}
\crefname{hypothesis}{Hypothesis}{Hypotheses}
\newsiamthm{claim}{Claim}

\title{Interpretable Approximation of High-Dimensional Data}

\author{Daniel Potts\thanks{Chemnitz University of Technology, Germany 
		(\email{potts@math.tu-chemnitz.de}, \url{http://www.tu-chemnitz.de/\string~potts/}).}
	\and Michael Schmischke\thanks{Chemnitz University of Technology, Germany 
		(\email{michael.schmischke@math.tu-chemnitz.de}, \url{http://www.tu-chemnitz.de/\string~mischmi/}).}
}

\usepackage{framed}
\colorlet{shadecolor}{blue!20}

\headers{Interpretable Approximation of High-Dimensional Data}{D. Potts, and M. Schmischke}

\ifpdf
\hypersetup{
  pdftitle={Interpretable Approximation of High-Dimensional Data},
  pdfauthor={D. Potts, and M. Schmischke}
}
\fi

\begin{document}

\maketitle

% REQUIRED
\begin{abstract}
	In this paper we apply the previously introduced approximation method based on the ANOVA (analysis of variance) decomposition and Grouped Transformations to synthetic and real data. The advantage of this method is the interpretability of the approximation, i.e., the ability to rank the importance of the attribute interactions or the variable couplings. Moreover, we are able to generate an attribute ranking to identify unimportant variables and reduce the dimensionality of the problem. We compare the method to other approaches on publicly available benchmark datasets.
\end{abstract}

% REQUIRED

% REQUIRED

\section{Introduction}

Building models for creating predictions based on empirical data is a current and active research topic with numerous applications. The amount of data being collected is ever-increasing resulting in high-dimensional datasets with corresponding regression or classification problems. There is a number of classical machine learning methods like support vector machines, neural networks, and decision trees, see e.g.~\cite{Hastie2013, Agg14, Bishop2016}, to deal with these problems. However, we simultaneously have the ever more important issue of interpretability of the models and therefore finding how the predictions come to pass. With this information one can e.g.~decide to not measure a certain value if it has little influence, in the case the measuring would be expensive, or use it to tune certain variables in order to achieve a desired outcome. While there is new research in the area of interpretability of the classical methods, see e.g.~\cite{SaWiMu17, MoSaMue18}, those models do not intrinsically allow for it. The method we apply in this paper does not only provide an alternative to classical machine learning methods, but also comes with a natural way to identify the importance and influence of attributes on the outcome.

In this paper we apply the approximation method presented in \cite{PoSc19a, PoSc19b} and propose ways in which to use them to analyze data and create predictions. The method is based on the idea of the analysis of variance (ANOVA) decomposition, cf.~\cite{CaMoOw97, RaAl99, LiOw06, KuSlWaWo09, Holtz11, Gu2013}, which allows us to approximate high-dimensional functions of a grouped structure when paired with the Grouped Transformations idea in \cite{BaPoSc}. The ANOVA decomposition uniquely decomposes a function into terms that correspond to variable couplings or variable interactions. The basis is the Lebesgue Hilbert space $\L_2$ for periodic functions or non-periodic functions defined over $[0,1]$ since here we have a complete orthonormal systems of functions and fast algorithms for multiplication in the Grouped Transformations, see \cite{PlPoStTa18, BaPoSc}.

Therefore, the only a-priori assumptions that we need are that the function we wish to approximate is either periodic or defined over a finite interval where it square-integrable as well as that the function can be well explained by limiting the variable interactions. This means that the significant part of the function is explained by letting only up to a number of variables interact simultaneously, see e.g.~\cite{CaMoOw97, KuSlWaWo09, DePeVo10, HaSchaeShiToTrWa21}, relating to the concept of the superposition dimension, see \cite{CaMoOw97, Ow19}. These assumptions are in general not very restricting and allows for a broad range of functions. Moreover, it has been theorized that most real word applications consist only of low-order interactions relating to sparsity-of-effects, cf.~\cite{Wu2011}, or the Pareto principle. 

Since the method gives us importance information on the variable couplings by using global sensitivity indices, cf.~\cite{So90, So01, LiOw06}, and sensitivity analysis, see \cite{saltelli08}, we can not only interpret this information, but moreover use it to improve the model by a number of techniques. Using attribute rankings we can remove an unimportant variable entirely and reduce the dimensionality of the problem. It is also possible to find that the representation of the data in the groups is sparse, i.e., some specific variable interactions does not influence the approximation (significantly) and can therefore be discarded. These techniques allow us to build an \textit{active set} of couplings that will in the end represent the model we use for predictions. This simultaneously gives us a control mechanism for the complexity of the model and combat overfitting. This technique is also related to low-dimensional structures and active subspace methods \cite{FoSchnaVy12, CoDoWa14, CoEfHoWa17} as well as random features \cite{RaRe08, ChiJiJa12, YaLiMaJiZh12, ChauSa19, HaSchaeShiToTrWa21}. The main difference to random features is that it draws weights or in our language indices/frequencies at random and uses a different optimization problem.

We apply the proposed approach to the Friedman functions, see \cite{MeLeHo03, BeGaMo09, BiDaLa11}, as an example of how it performs on synthetic data with Gaussian noise and compare our findings to previously obtained benchmark results in the same setting. Moreover, we test the method on real datasets for regression problems from the UCI database \cite{UCIr} and other sources. Each datasets provides a different challenge and we compare our results to previously obtained classical machine learning techniques. We observe very promising results and in many cases outperform previous benchmark experiments.

The paper is organizes as follows. In \cref{sec:anova} we reiterate on the previously in \cite{PoSc19a, PoSc19b, BaPoSc} introduced approximation method and how to use it for regression problems. Here, we discuss the required functional analytic foundations and introduce relevant complete orthonormal systems of functions. The procedure is explained in \cref{sec:procedure} and we also propose multiple techniques for model refinement and active set detection in \cref{sec:set_detection} that includes the computation of an attribute ranking. \cref{sec:synth} contains numerical experiments with synthetic data namely the Friedman functions. We recreate the setting from benchmark results and test our method under the same conditions. Moreover, in \cref{sec:real} we test the method on datasets from \cite{UCIr, TorgoWeb} and compare results to previous experiments. All numerical experiments are performed with the Julia package \cite{BaSchm20} and the code is available as a repository \cite{AttributeRankingCode}.
 
\section{Interpretable ANOVA Approximation}\label{sec:anova}

In this section we summarize on the interpretable ANOVA (Analysis of Variance) approximation method and the associated idea of grouped transformations, see \cite{PoSc19a, BaPoSc}. The approach was considered for periodic functions, but has since been expended to non-periodic functions in \cite{PoSc19b}. Therefore, we are able to utilize both types of approximations which brings advantages especially for real data.

We consider functions \begin{equation*}
	f \in \L_2(\dom^d) \coloneqq \left\{ \fun{f}{\dom^d}{\tar} \colon \norm{f}{\L_2(\dom^d)} \coloneqq \sqrt{\int_{\dom^d} \abs{f(\x)}^2 \d\x} < \infty \right\}
\end{equation*} with spatial dimension $d \in \N$ where $\dom$ is the torus $\T$ for 1-periodic functions and $D = [0,1]$ if we are non-periodic. We identify the torus with an interval of unit length specifically $\T \cong [-0.5,0.5)$. The function maps to the real numbers, i.e., $\tar = \R$, in the non-periodic case while $\tar = \C$ is possible for the periodic case. Furthermore, we have the scalar product \begin{equation*}
	\sp{f}{g} \coloneqq \int_{\dom^d} f(\x)\, g(\x) \d\x.
\end{equation*}

Now, let $\{ \phi_{\k} \}_{\k \in \Z^d}$ be a complete orthonormal system in the space $\L_2(\dom^d)$ with tensor product structure, i.e., we have a complete orthonormal system $\{ \eta_k \}_{k \in \Z}$ in  $\L_2(\dom)$ and $\phi_{\k}(\x) = \prod_{i=1}^{d} \eta_{k_i}(x_i)$. Then \begin{equation}\label{baserep}
	f(\x) = \sum_{\k \in \Z^d} \fc{\k}{f} \phi_{\k}(\x), \quad \fc{\k}{f} = \sp{f}{\phi_{\k}},
\end{equation} and through Parseval's identity $\norm{f}{\L_2(\dom^d)}^2 = \sum_{\k \in \Z^d} \abs{\fc{\k}{f}}^2$.

The classical ANOVA decomposition, cf.~\cite{CaMoOw97, RaAl99, LiOw06, Holtz11}, provides us with a unique decomposition in the frequency domain as shown in \cite{PoSc19a}. We denote the coordinate indices with $\D = \{ 1,2,\dots, d\}$ and subsets as bold small letters, e.g., $\u \subset \D$. An ANOVA term is defined as \begin{equation*}
	f_{\u}(\x) = f_{\u}(\x_{\u}) \coloneqq \sum_{\substack{\k \in \Z^d \\ \supp \k = \u}} \fc{\k}{f} \phi_{\k}(\x)
\end{equation*} with $\supp \k \coloneqq \{ s \in \D \colon k_s \neq 0 \}$ The function can then be uniquely decomposed as \begin{equation*}
f(\x) = \sum_{\u \subset \D} f_{\u}(\x)
\end{equation*} into $\abs{\mathcal{P}(\D)} = 2^d$ ANOVA terms where $\mathcal{P}(\D)$ is the potency set of $\D$. Here, the exponentially growing number of terms shows an expression of the curse of dimensionality in the decompositon. 

Crucial information to later achieve attribute rankings is the relative importance of an ANOVA terms $f_{\u}$ with respect to the function. In order to measure this we define the variance of a function $f$ as \begin{equation*}
	\va{f} \coloneqq \norm{f}{\L_2(\dom^d)}^2 - \abs{\fc{\b 0}{f}}^2 = \sum_{\k \in \Z^d\setminus\{\b 0\}} \abs{\fc{\k}{f}}^2.
\end{equation*} Note that $\va{f_{\u}} = \norm{f_{\u}}{\L_2(\dom^d)}^2$, $\u \subset \D$. The relative importance is now measured trough global sensitivity indices (GSI), see \cite{So90, So01, LiOw06}, defined as \begin{equation}\label{gsi}
	\gsi{\u}{f} \coloneqq \frac{\va{f_{\u}}}{\va{f}}.
\end{equation} This motivates the concept of effective dimensions. We focus specifically on the modified version of the superposition dimension as one notion of effective dimension. For a given $\alpha \in [0,1]$ it is defined as \begin{equation}\label{eq:ds}
	\mathrm{d}^{(\mathrm{sp})} \coloneqq \min \left\{ s \in \D \colon \sup_{\norm{f}{H(\dom^d)} \leq 1} \sum_{\au > s} \norm{f_{\u}}{\L_2(\dom^d)}^2 \leq 1-\alpha \right\}
\end{equation} for a function $f \in H(\dom^d)\subset \L_2(\dom^d)$. Here, $H(\dom^d)$ is $\L_2(\dom^d)$ or might be a Hilbert space that e.g.~characterizes smoothness trough the decay of the basis coefficients $\fc{\k}{f}$, cf.~\cite{PoSc19a, Ow19}.

Now, we still have the curse of dimensionality and need to find a way around it for efficient approximation. For this we take subsets of ANOVA terms $U \subset \mathcal{P}(\D)$ into account. These sets have to be downward closed, i.e., for every $\u \in U$ it holds that all subsets $\v \subset \u$ are also elements of $U$. We are then able to consider the truncated ANOVA decomposition \begin{equation*}
	\mathrm{T}_{U} f (\x) = \sum_{\u \in U} f_{\u}(\x). 
\end{equation*} A specific idea for the truncation comes from the superposition dimension $\mathrm{d}^{(\mathrm{sp})}$ in \eqref{eq:ds}. One might only take variable interactions into account that contain $d_s$ or less variables, i.e., the subset of ANOVA terms is \begin{equation*}
	U_{d_s} = \left\{ \u \subset \D\colon \au \leq d_s \right\}.
\end{equation*} Since $d_s$ can be any integer in $\D$ we call it superposition threshold. Note that $d_s$ can be equal to the superposition dimension, but this does not need to be the case. A well-known fact from learning theory is that the number of terms in $U_{d_s}$ grows only polynomially in $d$ for fixed $d_s < d$ which has reduced the curse of dimensionality.

\textbf{Why is the truncation through a superposition threshold $d_s$ a good idea?} Let us start this argument with the approximations of smooth functions that belong to some smoothness space $H(\dom^d) \subset \L_2(\dom^d)$. If the function has a low superposition dimension $\mathrm{d}^{(\mathrm{sp})}$ for $\alpha \in [0,1]$ the truncation by a low superposition threshold will be effective (in relation to $\alpha$). It is possible to characterize smoothness by the decay of the basis coefficients $\fc{\k}{f}$ and show upper bounds for the superposition dimension $\mathrm{d}^{(\mathrm{sp})}$ as in \cite{PoSc19a}. In fact there are types of smoothness that are proven to yield a low upper bound for the superposition dimension specifically dominating-mixed smoothness with POD (product and order-dependent) weights, cf.~\cite{KuSchwSl12, GrKuNi14, KuNu16, GrKuNu18, PoSc19a}. 

In terms of real data, the situation is much different. One thing in advance: For the complete generality of problems one cannot make the assumption that we have a low superposition dimension. However, there are many application scenarios where numerical experiments successfully showed that this is indeed the case, see e.g.~\cite{CaMoOw97}. Since we generally do not have a-priori information we work with low superposition thresholds $d_s$ for truncation and validate on a set of testdata. 

\subsection{Approximation Procedure}\label{sec:procedure}

We briefly discuss how an approximation is numerically obtained in a multiple step procedure and how we can interprete the results as well as create attribute rankings. In this section, we assume a given subset of ANOVA terms $U \subset \mathcal{P}(\D)$. How to obtain such a set will be discussed in \cref{sec:set_detection}. Moreover, we always assume that we have given scattered data in the form of a node set $\X = \{ \x_1, \x_2, \dots, \x_M \}\subset \dom^d$ and values $\y \in \tar^M$, $M \in \N$. We assume now that there is an $\L_2(\dom^d)$ function $f$ of form \eqref{baserep} with $f(\x_i) \approx y_i$ which we want to approximate.

The approximation procedure works as follows: We truncate $f$ to the terms in the set $U$ such that $f \approx \mathrm{T}_U f$. Since there are still infinitely many coefficients, we perform a truncation to partial sums with a finite support index set $I_{\u} \subset (\Z \setminus \{ \b 0  \})^{\au}$ for every ANOVA term $f_{\u}$, $\u \in U$, such that \begin{equation*}
	f_{\u}(\x) \approx \sum_{\k \in P_{\u} I_{\u}} \fc{\k}{f} \phi_{\k}(\x)
\end{equation*} and $P_{\u} I_{\u} = \{ \k \in \Z^d \colon \k_{\u} \in  I_{\u}, \k_{\uc} = \b 0 \}$. Taking the union $I(U) = \bigcup_{\u \in U} I_{\u}$, we have $f(\x) \approx \sum_{\k \in I(U)} \fc{\k}{f} \phi_{\k}(\x)$. The unknown coefficients $\fc{\k}{f}$ are now to be determined.

We aim to achieve this by solving the least-squares problem \begin{equation}\label{eq:minimization}
	\hat{\b f} = (\hat{f}_{\k})_{\k \in I(U)} = \argmin_{\hat{\b g} \in \tar^{\abs{I(U)}}} \norm{\y - \b{F}(\X,I(U)) \hat{\b g}}{2}
\end{equation} with the matrix $\b{F}(\X,I(U)) = (\phi_{\k}(\x))_{\k \in I(U), \x \in \X }$. If $\b{F}(\X,I(U))$ has full rank the problem has a unique solution and $\hat{f}_{\k} \approx \fc{\k}{f}$.  Then our approximation is \begin{equation*}
	f(\x) \approx S(\X,I(U)) f (\x) = \sum_{\k \in I(U)} \hat{f}_{\k} \phi_{\k}(\x).
\end{equation*} In general, if the oversampling $M/\abs{I(U)} > 1$, i.e., is large enough and the nodes are independent, one may assume full rank for $\b{F}(\X,I(U))$. 

\textbf{How can \eqref{eq:minimization} be efficiently solved?} In order to solve the minimization we employ the iterative LSQR solver \cite{PaSa82} which needs a method for efficient multiplication with $\b{F}(\X,I(U))$ and its adjoint $\b{F}^\ast(\X,I(U))$ in the periodic case, otherwise its transposed matrix. This is realized by the Grouped Transformation idea in \cite{BaPoSc} based on the NFFT or the NFCT, see \cite{KeKuPo09, PlPoStTa18}. We consider three distinct function systems in this paper for which we have a fast Grouped Transform. In the periodic case we use the Fourier system with \begin{equation}\label{eq:system:exp}
	 \phi^{\text{exp}}_{\k}(\x) = \e^{2\pi\i\k\cdot\x}.
\end{equation} For non-periodic functions we focus on the cosine system \begin{equation}\label{eq:system:cos}
	\phi^{\text{cos}}_{\k}(\x) = \sqrt{2}^{\abs{\supp \k}} \prod_{s \in \supp \k} \cos(\pi k_s x_s)
\end{equation} and the Chebyshev system \begin{equation}\label{eq:system:cheb}
\phi^{\text{cheb}}_{\k}(\x) = \sqrt{2}^{\abs{\supp \k}} \prod_{s \in \supp \k} \cos(k_i \arccos(2 x_s - 1)).
\end{equation} \begin{remark} 
	The Chebyshev system was considered in \cite{PoSc19b}: Although the theoretical approximation properties of this system are better, the nodes $\X$ optimally need to be distributed according to the Chebyshev probability measure and our experiments showed that this did not work well for problems with scattered data from real applications. In \cite{PoSc19b} we propose a way to circumvent this, by minimizing a weighted norm. However, this has negative effects on the condition of the system and results in the need for more data and/or iterations. Therefore, we opt to use the cosine system for scattered data. However, the situation is quite different if the nodes can be generated according a probability measure, e.g., in uncertainty quantification with PDE applications.
\end{remark} We also restrict ourselves to the usage of full-grid index sets, i.e., we use \begin{equation*}
	I_{\u}^{\text{per}} = \{ -N_{\u}/2, \dots, -1, 1, \dots, N_{\u}/2-1 \}
\end{equation*} in the periodic case and \begin{equation}\label{eq:set:nper}
	 I_{\u}^{\text{non-per}} = \{ 1,2,\dots,N_{\u}-1\}
\end{equation} for the non-periodic case with the even parameter $N_{\u} \in 2\N$. Note that the sets contain $N_{\u}-1$ elements in both cases.

It is possible to add different types of regularization to the problem \eqref{eq:minimization} which also allows to incorporate a-priori smoothness information. For details we refer to \cite{BaPoSc}. For our numerical experiments, we add $\ell_2$ regularization, i.e., solving \begin{equation}\label{regu}
	\argmin_{\hat{\b g} \in \tar^{\abs{I(U)}}} \norm{\y - \b{F}(\X,I(U)) \hat{\b g}}{2} + \lambda \norm{\hat{\b g}}{2}
\end{equation} with regularization parameter $\lambda > 0$.

\textbf{How can an attribute ranking be computed?} We use the global sensitivity indices $\gsi{\u}{S(\X,I(U)) f }$, $\u \in U$, from the approximation $S(\X,I(U)) f (\x)$ to compute approximations for the global sensitivity indices $\gsi{\u}{f}$ of the function $f$. Here, we do not consider the index to be a good approximation if the values are close together, but rather if there order is identical, i.e., we have \begin{equation*}
	\gsi{\u_1}{ f } \leq \gsi{\u_2}{f} \Longrightarrow \gsi{\u_1}{S(\X,I(U)) f } \leq \gsi{\u_2}{S(\X,I(U)) f } 
\end{equation*} for any pair $\u_1, \u_2 \in U$. 

If we are interested in how much one variable $i \in \D$ adds to the variance of the function, i.e., how important it is, we can compute the ranking score \begin{equation}\label{eq:ranking}
	r(i) = \frac{\sum_{\substack{\u \in U \\ i \in \u}} \abs{\{\v \in U \colon \au = \av, i \in \v\}}^{-1} \gsi{\u}{S(\X,I(U)) f }}{\sum_{\u \in U} \left(\sum_{i\in \u} \abs{\{\v \in U \colon \au = \av, i \in \v\}}^{-1}\right) \gsi{\u}{S(\X,I(U)) f } } .
\end{equation} The score attributes the contribution of every global sensitivity index to its variables weighted by the number of sets in that dimension and adds normalization such that $\sum_{i \in \D} r(i) = 1$. Dividing by the number of combinations in the same order is necessary since many terms with a low sensitivity index may otherwise yield a high score. Computing every score $r(i)$, $i \in \D$ provides an attribute ranking with respect to $U$ showing the percentage that every variable adds to the variance of the approximation. This allows for the conclusion that if we have a good approximation $S(\X,I(U)) f $, its attribute ranking will be close to the attribute ranking of the function $f$.

\subsection{Active Set Detection}\label{sec:set_detection}

In this section we describe how to obtain an active set of ANOVA terms $U$ for approximation. We are sill working with scattered data $\mathcal X = \{ \x_1, \x_2, \dots, \x_M \}\subset \dom^d$ and $\y \in \tar^M$, $M \in \N$. The values $\y$ may also contain noise. 

\textbf{Why does it make sense to reduce the number of ANOVA terms?}

The first step is to limit the variable interactions by a superposition threshold $d_s \in \D$ which may have been estimated by known smoothness properties (or different a-priori information) or set to a sensible value if nothing is known. Of course it is also possible to test and validate different values. We use the procedure described in \cref{sec:procedure} to obtain the approximation $S(\X,I(U_{d_s})) f$. For that we choose index sets $I_{\u}$ in an order-dependent way since there is no additional information available, i.e., $N_{\u_1} = N_{\u_2}$ for $\abs{\u_1} = \abs{\u_2}$. In order to get more consistence in the global sensitivity indices it is advisable to choose the index sets roughly of the same size, i.e., $(N_{\u_1}-1)^{\abs{\u_1}} \approx (N_{\u_2}-1)^{\abs{\u_2}}$ for $\u_1, \u_2 \in U_{d_s}$. 

From the approximation $S(\X,I(U_{d_s})) f$ we obtain the global sensitivity indices $\gsi{\u}{S(\X,I(U_{d_s})) f }$, $\u \in U$, and an attribute ranking $r(i)$, $i \in \D$, see \eqref{eq:ranking}. There are multiple ways to proceed from this point which we explain in the following.

\textbf{Removal of unimportant variables:} If the attribute ranking $r(i)$ shows variables that have very little to no influence on the variance of the function, then those variables may be removed entirely. Removing them leads to a reduction in the dimensionality of the problem and greatly simplifies the model function. 

\textbf{Active Set Thresholding:} Here, one chooses a threshold vector $\b\eps \in (0,1)^{d_s}$ and reduces the ANOVA terms to the set \begin{equation*}
	U(\b\eps) \coloneqq \left\{ \u \in U_{d_s} \colon \gsi{\u}{S(\X,I(U_{d_s})) f } > \eps_{\au}  \right\}.
\end{equation*} This set is not downward closed by definition, but if we set for all subsets $\v$ of the sets $\u \in U(\b\eps)$ with $\v \notin U(\b\eps)$ that $f_{\v} \equiv 0$ the condition is fulfilled. The parameter vector $\b\eps \in (0,1)^{d_s}$ allows control over how much of the variance may be sacrificed in order to simplify the model function.

\textbf{Incremental Expansion:} This method is advantageous if the model function is already very complex with a small superposition threshold $d_s$ which may occur if we are dealing with an especially large spatial dimension $d$. Here, we start with a small $d_s$, e.g., $d_s =1$ or $d_s = 2$. A reduction in the ANOVA terms can be performed by either of the two previous approaches. Now, one chooses a $\theta \in (0,1)$ and determines the subset \begin{equation*}
	\v \coloneqq \left\{ i \in \D \colon r(i) > \theta \right\}.
\end{equation*} If we assume that additional interactions of the important variables might also be significant to the variance, we may add interactions of size up to $n_{\v} \in \N$, $d_s < n_{\v} < d$. This translates to adding the set of terms \begin{equation*}
	U(\v, n_{\v}) \coloneqq \left\{ \u \in \mathcal{P}(\v) \colon d_s < \au \leq n_{\v} \right\}.
\end{equation*} This will be a beneficial way to improve the accuracy of the model if higher-order interactions play a role. However, if this method is used without reducing the complexity with any of the previous approaches, the overall complexity of the model will be higher.

In summary, one has to interpret the information obtained from the approximation $S(\X,I(U_{d_s})) f$ and choose the best performing methods based on this. Any combination of the previously mentioned approaches may lead to the optimal approximation as we will see in the numerical experiments. Moreover, it is possible and advisable to iterate this procedure multiple times, i.e., cross-validate it with the proposed active set detection steps in order to obtain the best result.

\section{Numerical Experiments with Synthetic Data}\label{sec:synth}

In this section we test our approximation and attribute ranking idea on synthetic data, i.e., we have a function $\fun{f}{\dom^d}{\R}$ and approximate it from artificially generated scattered data. Here, we focus on the non-periodic setting, i.e., $\dom = [0,1]$ and $\tar = \R$, and the cosine basis $\phi^{\text{cos}}_{\k}(\x)$, see \eqref{eq:system:cos}. Note that extensive tests for the periodic setting with the Fourier system \eqref{eq:system:exp} and the non-periodic Chebyshev system \eqref{eq:system:cheb} have been conducted in \cite{PoSc19a, PoSc19b, BaPoSc}. For a fixed number of nodes $M \in \N$ we create the node set $\mathcal{X} = \{\x_1,\x_2,\dots,\x_M\}\subset [0,1]^d$ by drawing uniform i.i.d.~nodes and evaluate the function $\y = (f(\x_i)+\eta)_{\i = 1}^M$. Moreover, the evaluations $\y$ may also contain noise $\eta$.

We are using the Friedmann functions, cf.~\cite{MeLeHo03, BeGaMo09, BiDaLa11} for our experiments. Our goal is to achieve the most accurate approximation which we will compare to the quality of different approaches known from the literature. In order to measure the quality of the approximation we take a second set of uniformly distributed i.i.d.~nodes $\X_{\text{test}} \subset [0,1]^d$, $\abs{\X_{\text{test}}} \in \N$, and choose the mean square error (MSE) as a measure of quality which is defined as \begin{equation}\label{mse}
	\mathrm{MSE}(f, \tilde{f}) = \frac{1}{\abs{\X_{\text{test}}}} \sum_{\x \in \X_{\text{test}}} \abs{f(\x)-\tilde{f}(\x)}^2.
\end{equation} Here, $\tilde{f}$ is the approximation or model for $f$ whose quality is to be measured.

In our experiments we will make us of regularization when solving problem \eqref{eq:minimization}, i.e., solving the modified problem \eqref{regu} which is described in more detail in \cite{BaPoSc}. Note that we rely solely on the $\ell_2$ variant of the regularization and the parameter will be denoted with $\lambda > 0$. The experiments have been conducted using the Julia package \cite{BaSchm20} and the code examples can be found online in \cite{AttributeRankingCode}.

\subsection{Friedmann Functions}

The Friedmann functions were used as benchmark examples in \cite{MeLeHo03} and have since become an often used example in the approximation of functions with scattered data, see e.g.~\cite{BeGaMo09, BiDaLa11}. We start by defining the three non-periodic Friedmann functions over $[0,1]^d$.

The first function 
\begin{equation*}
	\fun{f_1}{[0,1]^{10}}{\R}, f_1(\x) = 10\sin(\pi x_1 x_2)+20(x_3-0.5)^2+10 x_4 + 5 x_5
\end{equation*} has spatial dimension $10$. However, only five of the ten variables have any influence on the function which is the most important information we aim to find with our attribute ranking. Additionally, no more than two variables interact simultaneously, i.e., there will not be an error because of the ANOVA trunction with $U_{d_s}$ for a superposition threshold $d_s = 2$. In other words, setting $\alpha = 1$ yields $\mathrm{d}^{(\mathrm{sp})}  = 2$ in \eqref{eq:ds}. 

The second function 
\begin{equation*}
	\fun{f_2}{[0,1]^{4}}{\R}, \, f_2(\x) = \sqrt{ s_1^2(x_1) + \left( s_2(x_2) \cdot x_3 - \frac{1}{s_2(x_2) \cdot s_4(x_4)} \right)^2 }
\end{equation*} has spatial dimension $4$ and contains the variable scalings $s_1(x_1) = 100x_1$, $s_2(x_2) = 520\pi x_2 + 40\pi$, and $s_4(x_4) = 10x_4+1$. The scalings are necessary since we want to stay in the interval $[0,1]$ with each variable. As for the Friedmann 1 function, there are at most two variables interacting simultaneously, i.e., $\alpha = 1$ yields $\mathrm{d}^{(\mathrm{sp})}  = 2$ in \eqref{eq:ds} again.

The third and last Friedmann function is given by
\begin{equation*}
	\fun{f_3}{[0,1]^{4}}{\R}, \, f_3(\x) = \arctan\left( \frac{s_2(x_2) \cdot x_3 - (s_2(x_2) \cdot s_4(x_4))^{-1} }{s_1(x_1)} \right)
\end{equation*} again with spatial dimension $d=4$ and the same scalings $s_1$, $s_2$, and $s_4$ as before. Here, every term is (analytically) nonzero which means that entire function (without error) can only be reconstructed for a superposition threshold $d_s = 4$.

In order to compare our results to the experiments in \cite{MeLeHo03}, we choose to replicate the setting exactly: We use randomly generated sets $\X^{(i)} \subset [0,1]^{d}$ for each of the Friedman functions $i=1,2,3$ with uniformly distributed i.i.d.~nodes such that $M = \abs{\X^{(i)}} = 200$ for the model training. The dimensions are $d=10$ for Friedman 1, and $d=4$ for Friedman 2 and 3. Moreover, we add Gaussian noise $\eta_i$, $i=1,2,3$, to the function evaluations with a mean of zero and a variance of $\sigma_1 = 1$ for Friedman 1, $\sigma_2 = 125$ for Friedman 2, and $\sigma_3 = 0.1$ for Friedman 3, i.e., $\y^{(i)} = (f_i(\x)+\eta_i)_{\x \in \X^{(i)}}$. In order to validate the accuracy of the model, we use test sets of randomly drawn uniformly distributed i.i.d.~nodes $\X^{(i)}_{\mathrm{test}} \subset [0,1]^{d}$ for each of the Friedman functions $i=1,2,3$ with $M_{\mathrm{test}} = \abs{\X^{(i)}_{\mathrm{test}}} = 1000$. The function values are again evaluations with Gaussian noise such that $\y^{(i)}_{\mathrm{test}} = (f_i(\x)+\eta_i)_{\x \in \X^{(i)}_{\mathrm{test}}}$.

\cref{tab:friedman} contains the benchmark data from \cite{MeLeHo03} with a support vector machine (SVM), a linear model (lm), a neural network (mnet) and a random forest (rForst) as well as the results with our method (ANOVAapprox). In the following sections we discuss the detailed procedure on how to obtain the models for ANOVAapprox. Note that we have used the non-periodic cosine basis \eqref{eq:system:cos}.

\begin{table}[tbhp]
	\begin{center}
		\begin{tabular}{cccccc} 
			\toprule
			 & svm & lm & mnet & rForst & ANOVAapprox \\ 
			\midrule 
			Friedman 1 & 4.36 & 7.71 & 9.21 & 6.02 & \textbf{1.43} \\
			Friedman 2 ($\cdot \, 10^3$) & 18.13 & 36.15  & 19.61  & 21.50 &  \textbf{17.21}  \\
			Friedman 3 ($\cdot \, 10^{-3}$) & 23.15 & 45.42 & \textbf{18.12} & 22.21 & 20.69 \\
			\bottomrule
		\end{tabular}
		\caption{Mean squared errors (MSE) for different methods when approximating Friedman functions in \cite{MeLeHo03} compared to ANOVAapprox. The value for ANOVAapprox was obtained by training the model on 100 randomly generated training sets and validating them on 100 randomly generated test sets. All values are the medians of the experiment MSEs and the best value for every function is highlighted.}\label{tab:friedman}
	\end{center}
\end{table} 

The results show that the ANOVA approximation method is competitive to the other approaches and delivers the best MSE for Friedman 1 and 2 as well as a close second best MSE for Friedman 3. Note that a set with 200 datapoints is rather small and other experiments used significantly more data, but we aimed to stay in the exact setting of \cite{MeLeHo03}.

\subsubsection{Friedman 1}

The first Friedman function $f_1$ provides a good challenge for attribute ranking since it is a 10-dimensional function with only 5 variables that influence its value. Moreover, we only have as few as 200 nodes available for approximation. The only known information is the node set $\X^{(1)}$ and the noisy evaluations $\y^{(1)}$. The noise is Gaussian with zero mean and variance $\sigma_1 = 1$.

\begin{figure}[tbhp]
	\centering
	\begin{tikzpicture}
		\begin{axis}[ylabel={$r(i)$},
			xtick={1,2,3,4,5,6,7,8,9,10}]
			\addplot+[ycomb, OliveGreen, mark=star, mark options={OliveGreen, scale=2}] plot coordinates
			{( 1, 0.19912471528486084) ( 2, 0.2204563871364706) ( 3, 0.07647662884287301) ( 4, 0.3481366317440077) ( 5, 0.143281703162498)};
			\addplot+[ycomb, Dandelion, mark=triangle*, mark options={Dandelion}] plot coordinates
			{( 6, 0.004962632585347727) ( 7, 0.0015457063355586902) ( 8, 0.0016726700920864414) ( 9, 0.0011795248463308374) ( 10, 0.0031633999699659404)};
			\addplot[red,sharp plot,update limits=false, dashed, thick] 
			coordinates {(0,0.02) (11,0.02)};
		\end{axis}
	\end{tikzpicture}
	\caption{Attribute ranking of the Friedman 1 function using 200 nodes $\X^{(1)}$ and noisy evaluations $\y^{(1)}$ with $N_1 = 4$, $N_2 = 2$, regularization paramter $\lambda = 3$, and superposition threshold $d_s = 2$.}
	\label{fig:friedman1:ranking}
\end{figure}
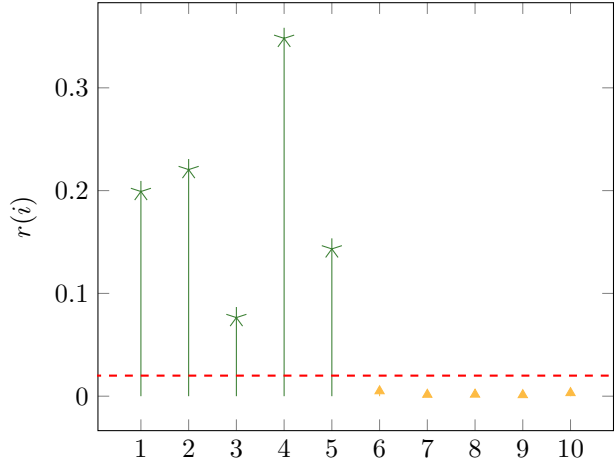

We begin by setting the superposition threshold to $d_s = 2$. Moreover, we choose for index sets $I_\emptyset = \{0\}$, $I_{\u} = \{1,\dots,N_1-1\}$, $N_1 \in \N$, for $\au = 1$, and $I_{\u} = \{1,\dots,N_2-1\}^2$, $N_2 \in \N$, for $\au = 2$. In \cref{fig:friedman1:ranking} we have computed an attribute ranking, see \eqref{eq:ranking}, for $f_1$ which clearly shows that the variables $x_6$ to $x_{10}$ are significantly less important than the others and indicate that we may remove them completely, i.e., the active set for approximation changes to \begin{equation*}
	U^{(\mathrm r)} = \{ \u \subset \{1,2,3,4,5\} \colon \au \leq 2 \}.
\end{equation*} Note that we have computed multiple attribute rankings and displayed the one where the corresponding approximation $S(\X^{(1)},I(U_2))f_1$, $\abs{I(U_2)} = 76$, achieved the best MSE on the test set $\X^{(1)}_{\mathrm{test}}$ of $4.99$, i.e., the closest model to the original function $f_1$.

We repeat our approximation process with $U^{(\mathrm r)}$ as active set of terms and consider the approximation $S(\X^{(1)},I(U^{(\mathrm r)}))f_1$, still with the parameters $N_1$, and $N_2$. Since $1 + \binom{5}{1} + \binom{5}{2} = 16$, we have $\abs{U^{(\mathrm r)}} = 16$ and therefore as many global sensitivity indices $\gsi{\u}{S(\X^{(1)},I(U^{(\mathrm r)}))f_1}$ to consider. \cref{fig:friedman1:gsi} shows the global sensitivity indices for the parameter constellation that yielded the best MSE of $2.50$ on the test set $\X^{(1)}_{\mathrm{test}}$. The important sets are \begin{equation*}
	U^\ast_{(1)} = \{ \emptyset \} \cup \{ \{1\}, \{2\}, \{3\}, \{4\}, \{5\}, \{1,2\} \}.
\end{equation*} while the sets in $U^{(\mathrm r)} \setminus U^\ast_{(1)}$ have a smaller global sensitivity index by a large margin and will be removed from the active set in the next step.

\begin{figure}[tbhp]
	\centering
	\begin{tikzpicture}
		\begin{axis}[ylabel={$\gsi{\u}{S(\X_1,I(U^{(\mathrm r)}))f_1}$}]
			\addplot+[ycomb, OliveGreen, mark=star, mark options={OliveGreen, scale=2}] plot coordinates
			{(1, 0.1876592548163349) (2, 0.18895123192742846) (3, 0.08570214013365907) (4, 0.31844276775944036) (5, 0.10787483341281738) (6, 0.06951342738035136)};
			\addplot+[ycomb, Dandelion, mark=triangle*, mark options={Dandelion}] plot coordinates
			{(7, 0.0047190491257465255) (8, 0.004827216396220904) (9, 0.002934525477033284) (10, 0.004171699660012712) (11, 0.0028313975850521255) (12, 0.003859858113446111) (13, 0.005467336342334572) (14, 0.007057542885383195) (15, 0.005987718984739137)};
			\addplot[red,sharp plot,update limits=false, dashed, thick] 
			coordinates {(0,2e-2) (16,2e-2)};
			\node at (axis cs:1,0.23) [anchor=north] {$\{1\}$};
			\node at (axis cs:2.3,0.2325) [anchor=north] {$\{2\}$};
			\node at (axis cs:3,0.13) [anchor=north] {$\{3\}$};
			\node at (axis cs:3.1,0.34) [anchor=north] {$\{4\}$};
			\node at (axis cs:5,0.15) [anchor=north] {$\{5\}$};
			\node at (axis cs:6.5,0.11) [anchor=north] {$\{1,2\}$};
		\end{axis}
	\end{tikzpicture}
	\caption{Global sensitivity indices $\gsi{\u}{S(\X^{(1)},I(U^{(\mathrm r)}))f_1}$, $\u \in U^{(\mathrm r)}$, with parameters $N_1= 6$, $N_2=4$, and $\lambda = 1$. $U^\ast_{(1)}$ with circles and complement with rectangles.}
	\label{fig:friedman1:gsi}
\end{figure}

We conclude the consideration of the Friedman 1 function with the final approximation $S(\X^{(1)},I(U^\ast_{(1)}))f_1$. The results for different $N_1$, and $N_2$ are displayed in \cref{tab:friedman1:approx}. The best MSEon the test set $\X^{(1)}_{\mathrm{test}}$ we were able to obtain is $1.36$. As displayed in \cref{tab:friedman}, the best MSE achieved by different methods using the same number of nodes and the same noise was $4.36$ by a support vector machine. Validating the model with 100 randomly generated training and test datasets yielded a median MSE of 1.43.

\begin{table}[tbhp]
	\begin{center}
		\begin{tabular}{cccc} 
			\toprule
			$N_1$ & $N_2$ & $\abs{ I(U^\ast_{(1)}) }$ & $\mathrm{MSE}$  \\ 
			\midrule 
			4 & 2 & 17 & 3.48 \\
			6 & 2 & 27 & 3.50 \\
			8 & 2 & 37 & 3.58 \\ \midrule 
			4 & 4 & 25 & 1.53 \\
			6 & 4 & 35 & 1.36 \\
			8 & 4 & 45 & 1.36 \\
			\bottomrule
		\end{tabular}
		\caption{Numerical experiments with the Friedman 1 function using 200 nodes $\X^{(1)}$ and noisy evaluations $\y^{(1)}$ with regularization paramter $\lambda = 1$ and superposition threshold $d_s = 2$. The $\mathrm{MSE} \coloneqq \mathrm{MSE}(f_1,S(\X^{(1)},I(U^\ast_{(1)}))f_1)$ was computed on the test set $\X^{(1)}_{\mathrm{test}}$ with $1000$ nodes.}\label{tab:friedman1:approx}
	\end{center}
\end{table}

\subsubsection{Friedman 2}

The second Friedman function $f_2$ is only four-dimensional with every dimension playing a role in the function. Therefore, we skip the attribute ranking and straightforward try to identify an active set of terms. For that we rely on the 200 nodes $\X^{(2)}$ and the evaluations $\y^{(2)}$ with Gaussian noise that has a mean of zero and a variance of $\sigma_2 = 125$.

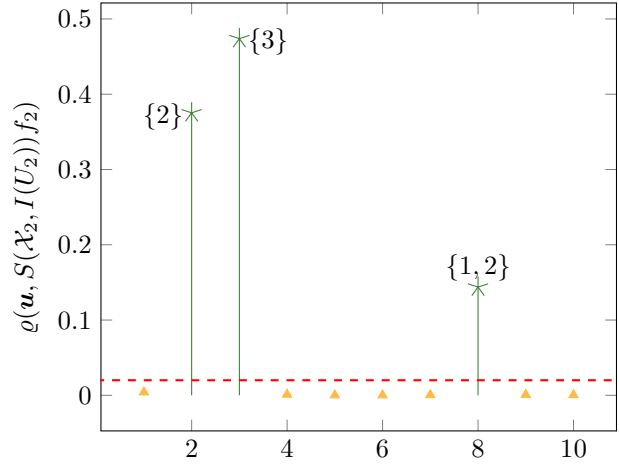
\begin{figure}[tbhp]
	\centering
	\begin{tikzpicture}
		\begin{axis}[ylabel={$\gsi{\u}{S(\X_2,I(U_2))f_2}$}]
			\addplot+[ycomb, OliveGreen, mark=star, mark options={OliveGreen, scale=2}] plot coordinates
			{ (2, 0.37532176225493) (3, 0.473872082392826) (8, 0.14380973570576244) };
			\addplot+[ycomb, Dandelion, mark=triangle*, mark options={Dandelion}] plot coordinates
			{(1, 0.004091329893730297) (4, 0.001013201673946832) (5, 7.938625636569526e-5) (6, 9.658382121862554e-5) (7, 0.00048563364189233466) (9, 0.000775505379832269) 
				(10, 0.00045477897949512604)};
			\addplot[red,sharp plot,update limits=false, dashed, thick] 
			coordinates {(0,2e-2) (11,2e-2)};
			\node at (axis cs:1.4,0.4) [anchor=north] {$\{2\}$};
			\node at (axis cs:3.6,0.50) [anchor=north] {$\{3\}$};
			\node at (axis cs:8,0.20) [anchor=north] {$\{1,2\}$};
		\end{axis}
	\end{tikzpicture}
	\caption{Global sensitivity indices $\gsi{\u}{S(\X^{(2)},I(U_2))f_2}$, $\u \in U_2$, with parameters $N_1= 4$, $N_2=2$, and $\lambda = 0$. $U^\ast_{(2)}$ with circles and complement with rectangles.}
	\label{fig:friedman2:gsi}
\end{figure}

As for the Friedman 1 function, we set the superposition threshold to $d_s = 2$. We also use full grid index sets $I_\emptyset = \{0\}$, $I_{\u} = \{1,\dots,N_1-1\}$, $N_1 \in \N$, for $\au = 1$, and $I_{\u} = \{1,\dots,N_2-1\}^2$, $N_2 \in \N$, for $\au = 2$ again. In \cref{fig:friedman2:gsi} we have visualized the global sensitivity indices $\gsi{\u}{S(\X^{(2)},I(U_2))f_2}$, $\u \in U_2$, for which $S(\X^{(2)},I(U_2))f_2$ yielded the best MSE of $17.37\cdot 10^{3}$ on the test set $\X^{(2)}_{\mathrm{test}}$. We are able to clearly identify the highlighted sets as important and use \begin{equation*}
	U^\ast_{(2)} = \{ \emptyset \} \cup \{ \{2\}, \{3\}, \{2,3\} \}.
\end{equation*} as active set going forward.

We proceed to approximate and show the results for different parameters in \cref{tab:friedman2:approx}. The best MSE achieved on the test set $\X^{(2)}_{\mathrm{test}}$ is $16.84 \cdot 10^3$ compared to $18.13 \cdot 10^3$ by a support vector machine, see \cref{tab:friedman}. Validating the model with 100 randomly generated training and test datasets yielded a median MSE of $17.21 \cdot 10^3$.

\begin{table}[tbhp]
	\begin{center}
		\begin{tabular}{cccc} 
			\toprule
			$N_1$ & $N_2$ & $\abs{ I(U^\ast_{(2)}) }$ & $\mathrm{MSE}$  \\ 
			\midrule 
			2 & 2 & 4 & $18.15 \cdot 10^{3}$ \\
			4 & 2 & 8 & $16.84 \cdot 10^{3}$ \\
			6 & 2 & 12 & $16.98 \cdot 10^{3}$ \\
			8 & 2 & 16 & $17.16 \cdot 10^{3}$ \\ \midrule 
			4 & 4 & 16 & $17.41 \cdot 10^{3}$ \\
			6 & 4 & 20 & $17.64 \cdot 10^{3}$ \\
			8 & 4 & 24 & $17.89 \cdot 10^{3}$ \\
			\bottomrule
		\end{tabular}
		\caption{Numerical experiments with the Friedman 2 function using 200 nodes $\X^{(2)}$ and noisy evaluations $\y^{(2)}$ with superposition threshold $d_s = 2$. The $\mathrm{MSE} \coloneqq \mathrm{MSE}(f_2,S(\X^{(2)},I(U^\ast_{(2)}))f_2)$ was computed on the test set $\X^{(2)}_{\mathrm{test}}$ of $1000$ nodes.}\label{tab:friedman2:approx}
	\end{center}
\end{table}

\subsubsection{Friedman 3}

The third Friedman function $f_3$ provides a challenge since all terms $f_{\u}$, $\u \subset \D$, are nonzero. Therefore, we are making a truncation error if the threshold $d_s$ is smaller than $d=4$. As before, we use only the 200 nodes in $\X^{(3)}$ and the evaluations $\y^{(3)}$ with Gaussian noise that has mean zero and variance $\sigma_3 = 0.1$. 

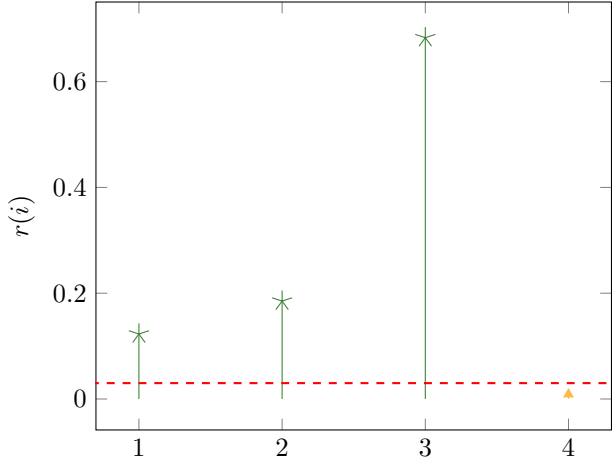
\begin{figure}[tbhp]
	\centering
	\begin{tikzpicture}
		\begin{axis}[ylabel={$r(i)$},
			xtick={1,2,3,4}]
			\addplot+[ycomb, OliveGreen, mark=star, mark options={OliveGreen, scale=2}] plot coordinates
			{( 1, 0.12288842541988297) ( 2, 0.18513640541594703) ( 3, 0.6831062171880993)};
			\addplot+[ycomb, Dandelion, mark=triangle*, mark options={Dandelion}] plot coordinates
			{( 4, 0.008868951976070823)};
			\addplot[red,sharp plot,update limits=false, dashed, thick] 
			coordinates {(0,0.03) (11,0.03)};
		\end{axis}
	\end{tikzpicture}
	\caption{Attribute ranking of the Friedman 3 function using 200 nodes $\X^{(3)}$ and noisy evaluations $\y^{(3)}$ with $N_1 = 10$, $N_2 = 2$, $N_3 = 2,$ regularization paramter $\lambda = 2$, and superposition threshold $d_s = 3$.}
	\label{fig:friedman3:gsi}
\end{figure}

At first we will use a superposition threshold of $d_s = 3$ to identify which ANOVA terms $f_{\u}$ with $\au \leq 3$ are important to the function. As before we rely on full grid index sets $I_\emptyset = \{0\}$, $I_{\u} = \{1,\dots,N_1-1\}$, $N_1 \in \N$, for $\au = 1$, $I_{\u} = \{1,\dots,N_2-1\}^2$, $N_2 \in \N$, for $\au = 2$, and $I_{\u} = \{1,\dots,N_3-1\}^3$, $N_3 \in \N$, for $\au = 3$. We have visualized the attribute ranking for the parameter choice that yielded the best MSE $2.18 \cdot 10^{-2}$ on the test set $\X^{(3)}_{\mathrm{test}}$ in \cref{fig:friedman3:gsi}. 

The ranking shows that the variable $x_4$ has very little influence on the approximation compared to the other 3 variables. This suggests that we are able to use \begin{equation*}
	U^{(\mathrm r)}_{d_s} = \{ \u \subset \{1,2,3\} \colon \au \leq d_s \}
\end{equation*} as active set. \cref{tab:friedman3:approx} shows the results for approximation of $f_3$ by $S(\X^{(3)},I(U^{(\mathrm r)}_{2}))f_3$ such that the superposition threshold $d_s = 2$ and $S(\X^{(3)},I(U^{(\mathrm r)}_{3}))f_3$ with $d_s =3$.

\begin{table}[tbhp]
	\begin{center}
		\begin{tabular}{ccccccc} 
			\toprule
			$N_1$ & $N_2$ & $N_3$ & $\abs{ I(U^{(\mathrm r)}_{2}) }$ & $\abs{ I(U^{(\mathrm r)}_{3}) }$ & $\mathrm{MSE}$ & $\overline{\mathrm{MSE}}$  \\ 
			\midrule 
			10 & 2 & 2 & 31 & 31 & $19.96 \cdot 10^{-3}$ & $20.25 \cdot 10^{-3}$ \\
			12 & 2 & 2 & 37 & 38 & $19.30 \cdot 10^{-3}$ & $19.63 \cdot 10^{-3}$ \\
			14 & 2 & 2 & 43 & 44 & $19.68 \cdot 10^{-3}$ & $20.06 \cdot 10^{-3}$ \\ \midrule
			10 & 4 & 2 & 55 & 56 & $22.74 \cdot 10^{-3}$ & $23.11 \cdot 10^{-3}$ \\
			12 & 4 & 2 & 61 & 62 & $22.12 \cdot 10^{-3}$ & $22.26 \cdot 10^{-3}$ \\
			14 & 4 & 2 & 67 & 68 & $24.26 \cdot 10^{-3}$ & $24.33 \cdot 10^{-3}$ \\
			\bottomrule
		\end{tabular}
		\caption{Numerical experiments with the Friedman 3 function using 200 nodes $\X^{(3)}$ and noisy evaluations $\y^{(3)}$. The $\mathrm{MSE} \coloneqq \mathrm{MSE}(f_3,S(\X_3,I(U^{(\mathrm r)}_{2}))f_3)$ and $\overline{\mathrm{MSE}} \coloneqq \mathrm{MSE}(f_3,S(\X_3,I(U^{(\mathrm r)}_{3}))f_3)$ were computed on the test set $\X^{(3)}_{\mathrm{test}}$ of $1000$ nodes.}\label{tab:friedman3:approx}
	\end{center}
\end{table}

We achieve a best MSE of $19.3 \cdot 10^{-3}$ on the test set $\X^{(3)}_{\mathrm{test}}$ with active set $U^{(\mathrm r)}_{2}$, i.e., we set the superposition threshold to $d_s = 2$ and do not need the three-dimensional term $\{1,2,3\}$. For comparison we find an MSE of $18.12 \cdot 10^{-3}$ as the best result in \cite{MeLeHo03}, cf.~\cref{tab:friedman}. Validating the model with 100 randomly generated training and test datasets yielded a median MSE of $18.12 \cdot 10^{-3}$.

%\subsection{DNN}
%
%\begin{equation}
%	\fun{g}{[0,1]^{50}}{\R}, \, g(\x) = \frac{1}{1+0.5\sum_{s=1}^{d} x_j j^{-2.5}}
%\end{equation}

\section{Numerical Experiments with Real Data}\label{sec:real}

In this section we test the ANOVAapprox method on datasets from real applications, i.e., we get a set of nodes $\X \subset [0,1]^d$ with $\abs{X} = M \in \N$ and noisy evaluations $\y \in \R^M$. Note that the data is not in $[0,1]$ in general, but we can achieve this trough min-max-normalization in a pre-processing step. Moreover, we have to decide how to split $\X$ into two parts, $\X_{\mathrm{train}}$ which we use for solving the optimization problem, i.e., training our model and obtaining the basis coefficients, and $\X_{\mathrm{test}}$ for validating our method and computing the error. Moreover, we focus on using the non-periodic cosine basis \eqref{eq:system:cos} and corresponding index sets $I_{\u}^{\text{non-per}}$, see \eqref{eq:set:nper}, for the ANOVA terms $f_{\u}$. Here, $N_{\u} \in 2\N$ is the associated bandwidth parameter that we always choose order-dependent, i.e., $N_{\u_1} = N_{\u_2} = N_{\abs{\u_1}}$ for $\abs{\u_1} = \abs{\u_2}$. As for the synthetic data, we make use of the $\ell_2$ regularization proposed in \cite{BaPoSc}, cf.~\eqref{regu}. The experiments have been conducted using the Julia package \cite{BaSchm20} and the code examples can be found online in \cite{AttributeRankingCode}.

\cref{tab:datasets} provides an overview of the data we use and where we obtained it. The datasets are all well-known and have been used for regression benchmarking in the past. We will not provide an in-depth description of the precise approximation steps for every dataset, but rather give a summary and compare the results in the end. Note that we use the root mean square error as a quality measure which describes the square root of the MSE \eqref{mse}, i.e., \begin{equation*}
	\mathrm{RMSE}(f, \tilde{f}) \coloneqq \sqrt{\mathrm{MSE}(f, \tilde{f})}.
\end{equation*} Moreover, for the \textit{Airfoil Self-Noise} problem we use the relative error \begin{equation*} 
	\sqrt{\frac{\sum_{\x \in \X_{\text{test}}} \abs{f(\x)-\tilde{f}(\x)}^2}{\sum_{\x \in \X_{\text{test}}} \abs{f(\x)}^2}}
\end{equation*} in order to compare our results to \cite{HaSchaeShiToTrWa21}.

\begin{table}[tbhp]
	\begin{center}
		\begin{tabular}{cccc} 
			\toprule
			Name & dimension & datapoints & references  \\ 
			\midrule 
			ENC & 8 & 768 & \cite{GoPaTh20, UCIr}   \\
			ENH & 8 & 768 & \cite{GoPaTh20, UCIr} \\
			Airfoil Self-Noise (ASN) & 5 & 1503 & \cite{HaSchaeShiToTrWa21, UCIr} \\
			California Housing (CH) & 8 & 20640 & \cite{KoMa15, TorgoWeb} \\
			Ailerons & 40 & 13750 & \cite{KoMa15, TorgoWeb} \\
			\bottomrule
		\end{tabular}
		\caption{Real datasets for benchmarking the ANOVAapprox method with sources.}\label{tab:datasets}
	\end{center}
\end{table}

\begin{figure}[tbhp]
	\centering
	\subfloat[ENC]{\label{fig:ar:enc}
		\begin{tikzpicture}[scale=0.75]
		\begin{axis}[ylabel={$r(i)$},
			xtick={1,2,3,4,5,6,7,8}]
			\addplot+[ycomb, OliveGreen, mark=star, mark options={OliveGreen, scale=2}] plot coordinates
			{( 5, 0.38873692846545005 )};
			\addplot+[ycomb, Dandelion, mark=triangle*, mark options={Dandelion}] plot coordinates
			{( 1, 0.06436798328098758 )( 2, 0.06656061627766091 )( 3, 0.08831828428879317 )( 4, 0.10226735420264628 )( 6, 0.10313566555893836 )( 7, 0.11600288763058085 )( 8, 0.07061028029494285 )};
		\end{axis}
	\end{tikzpicture}
	}
	\subfloat[ENH]{\label{fig:ar:enh}
	\begin{tikzpicture}[scale=0.75]
		\begin{axis}[ylabel={$r(i)$},
			xtick={1,2,3,4,5,6,7,8}]
			\addplot+[ycomb, OliveGreen, mark=star, mark options={OliveGreen, scale=2}] plot coordinates
			{( 5, 0.39556541656813937 )};
			\addplot+[ycomb, Dandelion, mark=triangle*, mark options={Dandelion}] plot coordinates
			{( 1, 0.07048005967684907 )( 2, 0.07022483059170234 )( 3, 0.08215990836456345 )( 4, 0.09648416505910608 )( 6, 0.08186593909510438 )( 7, 0.1437532788397012 )( 8, 0.05946640180483429 )		};
		\end{axis}
	\end{tikzpicture}
} \\ 	\subfloat[Airfoil Self-Noise]{\label{fig:ar:asn}
\begin{tikzpicture}[scale=0.75]
	\begin{axis}[ylabel={$r(i)$},
		xtick={1,2,3,4,5}]
		\addplot+[ycomb, OliveGreen, mark=star, mark options={OliveGreen, scale=2}] plot coordinates
		{( 3, 0.3006493615647468 )( 4, 0.43597531448775506 )};
		\addplot+[ycomb, Dandelion, mark=triangle*, mark options={Dandelion}] plot coordinates
		{( 1, 0.09758573144700278 )( 2, 0.09596580906362208 )( 5, 0.06982378343687337 )};
	\end{axis}
\end{tikzpicture}
}
\subfloat[California Housing]{\label{fig:ar:ch}
\begin{tikzpicture}[scale=0.75]
	\begin{axis}[ylabel={$r(i)$},
		xtick={1,2,3,4,5,6,7,8},
		ytick={0.1,0.2}]
		\addplot+[ycomb, OliveGreen, mark=star, mark options={OliveGreen, scale=2}] plot coordinates
		{( 1, 0.2278911676073449 )( 2, 0.2370023358813096 )( 6, 0.17285401586190402 )};
		\addplot+[ycomb, Dandelion, mark=triangle*, mark options={Dandelion}] plot coordinates
		{( 3, 0.050080876920213305 )( 4, 0.07520085724213338 )( 5, 0.07642716286668874 )( 7, 0.08933869139608512 )( 8, 0.07120489222432098 )		};
	\end{axis}
\end{tikzpicture}
}\\ 	\subfloat[Ailerons]{\label{fig:ar:ail}
\begin{tikzpicture}[scale=0.75]
	\begin{axis}[ylabel={$r(i)$},
		xtick={1,3,5,7,9,11},
		ytick={0.1,0.15,0.2,0.05}]
		\addplot+[ycomb, OliveGreen, mark=star, mark options={OliveGreen, scale=2}] plot coordinates
		{( 1, 0.2121482106725719 )( 2, 0.16501573294381774 )( 8, 0.17180623894806166 )};
		\addplot+[ycomb, Dandelion, mark=triangle*, mark options={Dandelion}] plot coordinates
		{( 3, 0.03266455895670102 )( 4, 0.05457947543336142 )( 5, 0.007048243946259695 )( 6, 0.057563386513982524 )( 7, 0.057563386513982524 )( 9, 0.11087793118774388 )( 10, 0.07316944836953507 )( 11, 0.057563386513982524 )};
	\end{axis}
\end{tikzpicture}
}
	\caption{Attribute ranking for the datasets from \cref{tab:datasets}.}
	\label{fig:ar}
\end{figure}
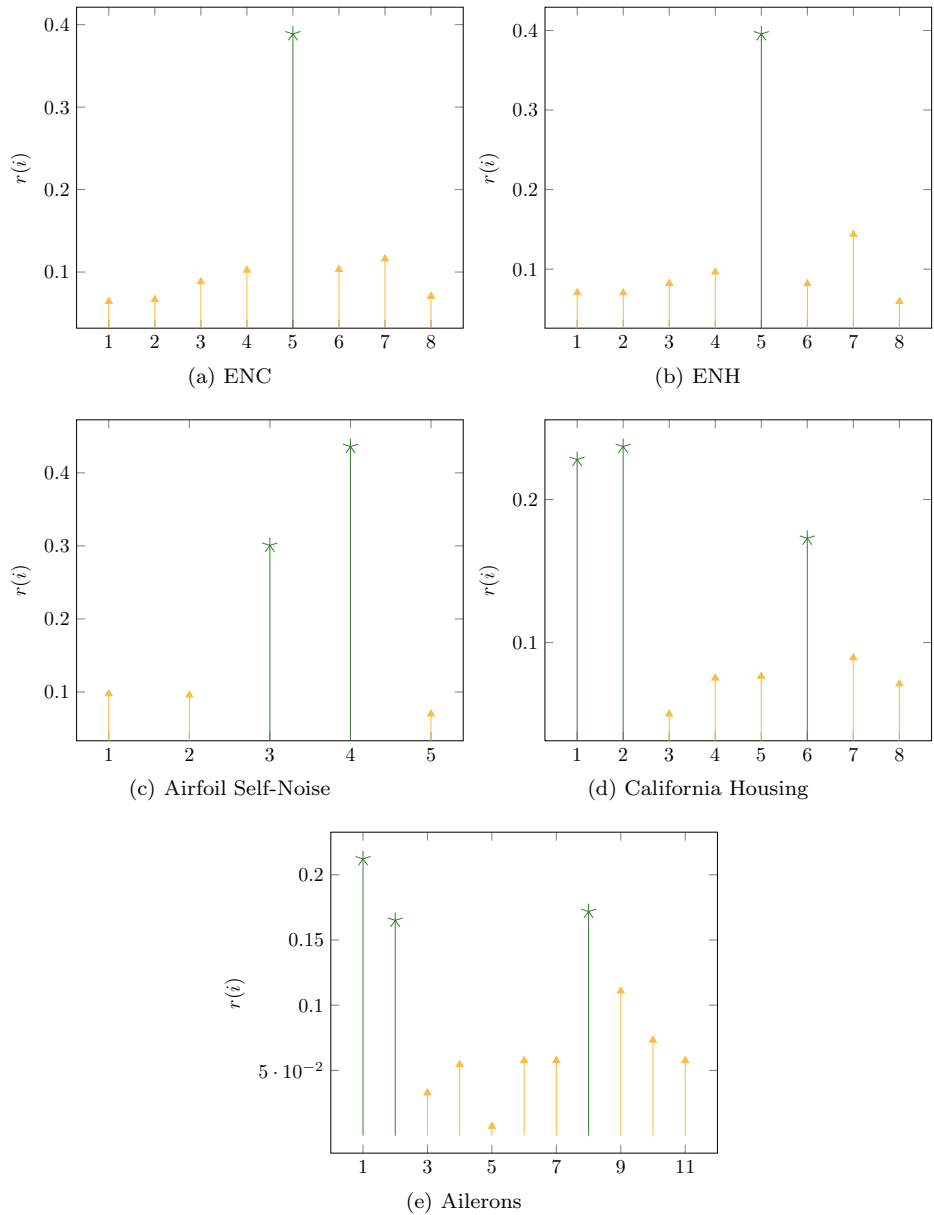

\cref{tab:realdata} shows the results we obtained with the ANOVA approximation approach compared to other methods. For the energy efficiency problems \textit{ENC} and \textit{ENH} we compare our results to \cite{GoPaTh20} where different classical machine learning and ensemble methods were tested on the same data. Our results outperform even the ensemble methods when comparing the RMSE. The problem of \textit{Airfoil Self-Noise} was considered as an example in \cite{HaSchaeShiToTrWa21} for the newly proposed method of sparse random features. Our obtained model was able to achieve a slightly better result by roughly one percent, see \cref{tab:realdata}. The remaining problems \textit{California Housing} and \textit{Ailerons} were considered as benchmark examples for multithreaded local learning regularization networks in \cite{GoPaTh20} where we are also able to achieve a lower RMSE. Note that we have tried to replicate the setting for every dataset, i.e., using the same percentages for training and testing as well as validating our model on 100 random splits. Moreover, \cref{fig:ar} contains the attribute rankings for each of the 5 models showing the importance of the different attributes for the datasets.

\begin{table}[tbhp]
	\begin{center}
		\begin{tabular}{cccc} 
			\toprule
			dataset & error (type) & method (reference) & ANOVAapprox \\ 
			\midrule 
			ENC & 1.79 (RMSE) & Gradient Boosting Machine (\cite{GoPaTh20}) & 1.49 \\
			ENH & 0.48 (RMSE) & Random Forest (\cite{GoPaTh20}) & 0.44 \\
			ASN & 0.0277 (relative) & Sparse Random Features (\cite{HaSchaeShiToTrWa21}) & 0.0161 \\
			CH & 0.11450 (RMSE) & Local Learning Reg.~NN (\cite{KoMa15}) & 0.10899 \\
			Ailerons & 0.04601 (RMSE) & Local Learning Reg.~NN (\cite{KoMa15}) & 0.04569 \\
			\bottomrule
		\end{tabular}
		\caption{Result comparison for different datasets and approaches. The models for ANOVAapprox where validated using 100 random splits of training and test set. More details are discussed in the corresponding subsection of \cref{sec:real}. The ANOVAapprox error is compared to the best error found in the mentioned source together with the method used therein.}\label{tab:realdata}
	\end{center}
\end{table} 

\subsection{Energy Efficiency}

The dataset describes the energy efficiency of houses by $8$ attributes and two values to predict, the cooling load and the heating load. Therefore, we have two problems ENC (8 attributes, 1 continuous value to predict) for the cooling load and ENH (8 attributes, 1 continuous value to predict) for the heating load. The dataset contains 768 samples which we split 70\% for the training set $\X_{\mathrm{train}}$ and 30\% for the test set $\X_{\mathrm{test}}$. The nodes $\X$ are normalized into $[0,1]$. 

First, we consider the ENC problem and start by setting the superposition threshold $d_s = 2$ and analyze the global sensitivity indices in order to remove unimportant sets. Experiments showed that removing sets or terms with a global sensitivity index (GSI) of less than $0.002$ yielded the best result. This leads to an active set $U_{\text{ENC}}^\ast \subset \mathcal P(\D)$ with $22$ terms. The resulting model gives a median RMSE of $1.49$ for 100 random splits into training and test set.

For the ENH problem we proceed in a similar fashion. We set the superposition threshold to $d_s = 2$ and then analyze the GSI of our model. Here, we choose the active set $U_{\text{ENH}}^\ast \subset \mathcal P(\D)$ consisting of all ANOVA terms with a GSI larger than $0.001$ such that $\abs{U_{\text{ENH}}^\ast} = 28$. As a result we obtain a model with a median RMSE of $0.44$ for 100 random splits into training and test set.

The optimal order-dependent bandwidths parameters $N_1, N_2 \in \N$ for both problems were computed using cross-validation. \cref{fig:ar:enc} and \cref{fig:ar:enh} show attribute rankings for our obtained models. We notice that the attribute 5, i.e., the overall height of the building, is especially important for the prediction in both problems.

\subsection{Airfoil Self-Noise}

This datasets originates from the NASA and contains data about NACA airfoils for different wind tunnel speeds and angles of attack. We aim to find a model that is able to predict the scaled sound pressure level of the self-noise in decibels (continuous), see \cite{UCIr}. The data contains 5 attributes and 1503 nodes. We perform a random split with 80\% for the training set $\X_{\mathrm{train}}$ and 20\% for the test set $\X_{\mathrm{test}}$. Since this dataset has recently been used in \cite{HaSchaeShiToTrWa21} for experiments with sparse random features, we choose the same split to compare the results. Note that the nodes were normalized into $[0,1]$.

An analysis of the global sensitivity indices for the superposition threshold $d_s = 2$ shows that there is only one unimportant term with a GSI less than $0.001$ that is to be removed. Therefore, we have an active set $U_{\text{ASN}}^\ast \subset \mathcal P(\D)$ with $\abs{U_{\text{ASN}}^\ast} = 14$ and need to use cross-validation in order to determine the optimal order-dependent bandwidths parameters $N_1, N_2 \in \N$. The obtained model was validated on 100 random 80/20 splits into training and test data yielding a median relative error of $1.61$\%. In \cref{fig:ar:asn} we have visualized the attribute ranking for our model. It shows that attributes 3 and 4, i.e., the chord length and the free-stream velocity have a large influence on the predictions.

\subsection{California Housing}

The datasets describes the prices for houses in California using data about the block groups from the 1990 census. Using 8 attributes and a set of 20460 cases, we aim to predict the median house price for the area. Since we want to compare our results to \cite{KoMa15}, we have split the data in 50\% for the training set $\X_{\mathrm{train}}$ and another 50\% for the test set $\X_{\mathrm{test}}$. The nodes \textbf{as well as the evaluations} were normalized into $[0,1]$. The normalization of the evaluations is replicated from \cite{KoMa15}.

We used a superposition threshold of $d_s = 2$ and analyzed the GSIs of the ANOVA terms. This lead to an active set $U_{\text{CH}}^\ast \subset \mathcal P(\D)$ with $\abs{U_{\text{CH}}^\ast} = 21$  terms. The bandwidth parameters $N_1, N_2 \in \N$ were then computed using cross-validation. The model was subsequently validated on 100 random 50/50 splits of the training and test data which yielded a median RMSE of $0.10899$. \cref{fig:ar:ch} shows the attribute ranking for the obtained model hinting that the variables 1, 2, and 6, i.e., the geographical location and the population count, are most important for the prediction. It also evident from the GSI that the ANOVA term $f_{\{1,2\}}$ has significant importance which makes sense since variable 1 is the longitude and variable 2 the latitude and together they represent the geographical location.

\subsection{Ailerons}

The Ailerons dataset describes the control problem of flying a F16 aircraft. The attributes describe the status of the aircraft while we aim to predict the control action on its ailerons. We have 40 attributes and 13750 samples. In order to replicate the setting in \cite{KoMa15}, we have split the data in 50\% for the training set $\X_{\mathrm{train}}$ and another 50\% for the test set $\X_{\mathrm{test}}$. The nodes \textbf{as well as the evaluations} were normalized into $[0,1]$. The normalization of the evaluations is replicated from \cite{KoMa15}.

We started to consider an attribute ranking for superposition threshold $d_s = 1$ in order to check if some variables have little influence and can be omitted for the model. This lead us to eliminate $29$ variables with a small contribution. We determined this number through cross-validation. Afterwards, we proceeded with the $11$ active variables and $d_s = 2$. A sensitivity analysis leads to the elimination of more terms leading to an active set $U_{\text{Ail}}^\ast \subset \mathcal P(\D)$ with $\abs{U_{\text{Ail}}^\ast} = 43$  terms. A validation of our model on 100 random 50/50 splits into training and test data has yielded a median RMSE of $0.04569$. In \ref{fig:ar:ail} we have visualized the attribute ranking for our model showing that variables 1,2, and 8 are important. They correspond to the variables 7, 3, and 30 of the original problem.

\section{Conclusion}

Numerical experiments with synthetic and real data showed that the proposed approach for approximation using ANOVA and Grouped Transformations, see \cite{PoSc19a, PoSc19b, BaPoSc}, is a competitive method in the approximation of high-dimensional data outperforming even ensemble machine learning methods in our experiments. Moreover, it delivers additional evidence for the fact that in applications we are able to assume that functions consist of (mostly) low-order interactions or are at least explained well by them. Since the method allows intrinsically for interpretation, we are able to produce an attribute ranking that shows how much different attributes influence the predictions. This can also be used to improve the model by removing unimportant variables or variable interactions entirely. Finally, we have proposed and applied multiple methods for the detection of an active set of ANOVA terms.

\section*{Acknowledgments}
We thank our colleagues in the research group SAlE for valuable discussions on the contents of this paper. Daniel Potts acknowledges funding by Deutsche Forschungsgemeinschaft (German Research Foundation) -- Project--ID 416228727 -- SFB 1410. Michael Schmischke is supported by the BMBF grant 01$|$S20053A. 

\bibliographystyle{siamplain}
\bibliography{../refs/references.bib}
\end{document}